\documentclass[conference]{IEEEtran}
\IEEEoverridecommandlockouts
\usepackage{cite}
\usepackage{amsmath,amssymb,amsfonts}
\usepackage{algorithmic}
\usepackage{graphicx}
\usepackage{textcomp}
\usepackage{xcolor}
\usepackage[raggedright]{sidecap}
\usepackage[singlelinecheck=false,justification=justified]{caption}
\usepackage{booktabs}
\usepackage{multirow}

\usepackage{caption}
\usepackage{subcaption}

\usepackage{sidecap}
\sidecaptionvpos{figure}{t}
\usepackage[colorlinks]{hyperref}

\newcommand\blfootnote[1]{%
  \begingroup
  \renewcommand\thefootnote{}\footnote{#1}%
  \addtocounter{footnote}{-1}%
  \endgroup
}

\newif\ifworkinprogress
\workinprogresstrue

\ifworkinprogress
	\newcommand{\ms}[1]{\textcolor{blue}{{[Markus] #1}}}
	\newcommand{\mm}[1]{\textcolor{olive}{{[Marta] #1}}}
	\newcommand{\sn}[1]{\textcolor{green}{{[Shah] #1}}}
      
\else
    \newcommand{\ms}[1]{}
    \newcommand{\mm}[1]{}
    \newcommand{\sn}[1]{}
    
\fi

\makeatletter
\let\@oldmaketitle\@maketitle
\renewcommand{\@maketitle}{\@oldmaketitle
  \vspace{-2pt}
  \setcounter{figure}{0}
  \centering\includegraphics[width=0.99\linewidth]{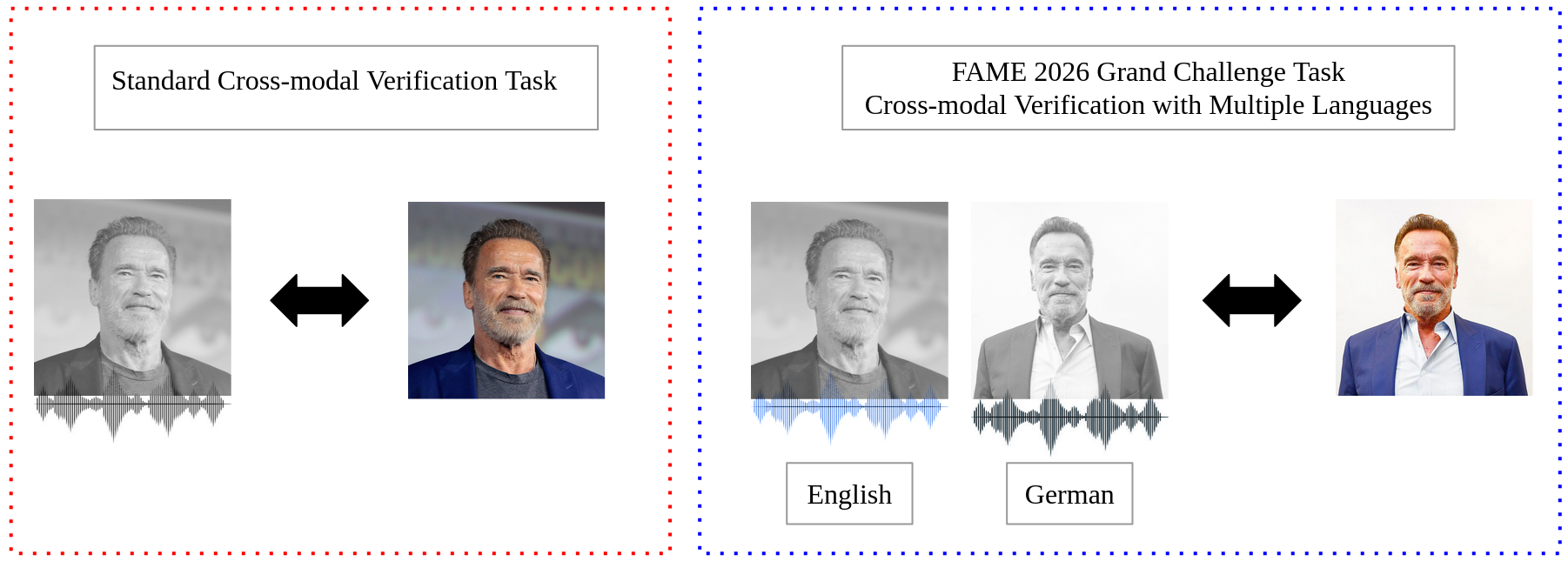}
  \vspace{-6pt}
  \captionof{figure}{(Left) Face-Voice association is established with a cross-modal verification task~\cite{nagrani2018learnable}. (Right) The FAME 2026 Challenge extends the task to analyze the impact of multiple languages.}
  \label{fig:verification+protocol}
  \vspace{0pt}
 }
\makeatother

\def\BibTeX{{\rm B\kern-.05em{\sc i\kern-.025em b}\kern-.08em
    T\kern-.1667em\lower.7ex\hbox{E}\kern-.125emX}}
\begin{document}

\title{\huge Face-voice Association in Multilingual Environments (FAME)  2026 Challenge Evaluation
Plan}

\author{ Marta Moscati$^{1}$\textsuperscript{\textdagger}, Ahmed Abdullah$^{2}$\textsuperscript{\textdagger}, Muhammad Saad Saeed$^{3}$\textsuperscript{\textdagger}, Shah Nawaz$^{1}$\textsuperscript{\textdagger}, Rohan Kumar Das$^{4}$\textsuperscript{\textdagger}, \\ Muhammad Zaigham Zaheer$^{5}$,  Junaid Mir$^{6}$, Muhammad Haroon Yousaf$^{6}$, Khalid Malik$^{3}$, Markus Schedl$^{6,7}$  \\
$^{1}$Johannes Kepler University Linz, Austria, 
$^{2}$National University of Computer and Emerging Sciences, Pakistan\\
$^{3}$University of Michigan, USA, 
$^{4}$Fortemedia Singapore, Singapore \\
$^{5}$Mohamed bin Zayed University of Artificial Intelligence, United Arab Emirates \\
$^{6}$University of Engineering and Technology Taxila, Pakistan,\\
$^{7}$Human-centered AI Group, AI Lab, Linz Institute of Technology, Austria \\
\tt mavceleb@gmail.com
}


\maketitle




\begin{abstract}
The advancements of technology have led to the use of multimodal systems in various real-world applications. Among them, audio-visual systems are among the most widely used multimodal systems. In the recent years, associating  face and voice of a person has gained attention due to the presence of unique correlation between them. The Face-voice Association in Multilingual Environments (FAME) 2026 Challenge focuses on exploring face-voice association under the unique condition of a multilingual scenario. This condition is inspired from the fact that half of the world's population is bilingual and most often people communicate under multilingual scenarios. The challenge uses a dataset named Multilingual Audio-Visual (MAV-Celeb) for exploring face-voice association in multilingual environments. This report provides the details of the challenge, dataset, baseline models, and task details for the FAME Challenge.

\end{abstract}

\section{Introduction}
\blfootnote{\textsuperscript{\textdagger}Equal Contribution.}
The face and voice of a person have unique characteristics and they are often used as biometric measures for person authentication, either as a unimodal or multimodal input~\cite{jain2004introduction,ross2004multimodal}. The strong correlation that humans establish in their perception of faces and voices of different people has inspired the development of tools for automated face-voice association
~\cite{nawaz2019deep,tao20b_interspeech,zheng2021adversarial,shah2023speaker,hannan2025paeff}. Though previous works have established association between faces and voices, none of these approaches investigated the effect of multiple languages on this task. 
Since half of the world's population is bilingual and we are often communicating in multilingual scenarios, it is important to explore the influence of language on associating faces with voices. The goal of the FAME challenge, as also summarized in Fig.~\ref{fig:verification+protocol}, is to analyze the impact of multilingual scenarios on face-voice association tasks.


\section{Grand Challenge Objectives}
The goal of the Face-voice Association in Multilingual Environments (FAME) challenge, as also summarized in Fig.~\ref{fig:verification+protocol}, is to analyze the impact of multilingual scenarios on face-voice association tasks.  More concretely, the aim of the challenge is to promote research and development on the following aspects of face-voice association: 
\begin{itemize}
    \item Explore the influence of language on the performance of face-voice association algorithms 
    \item Promote the development novel face-voice association algorithms that are effective in multilingual scenarios
\end{itemize}

\begin{figure}[t]
    \centering
    \includegraphics[width=0.99\linewidth]{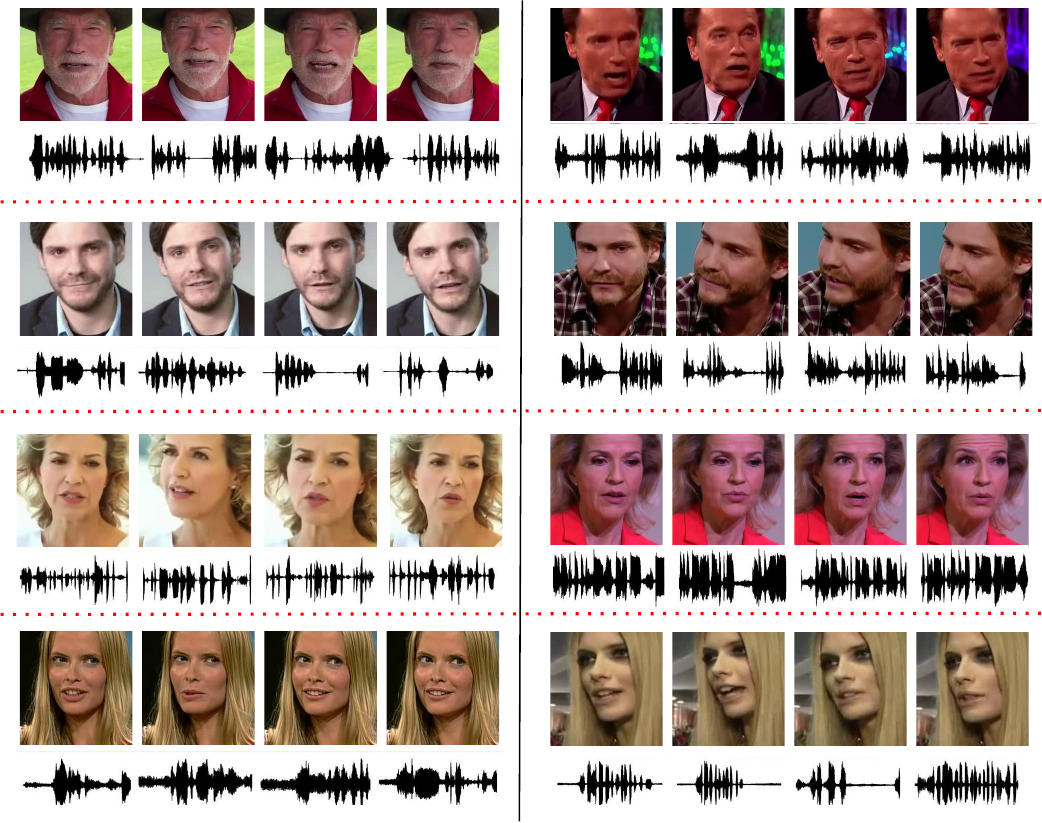}
    \caption{Audio-visual samples selected from the MAV-Celeb dataset. The visual data contains different variations such as pose, lighting condition, and motion. The left block shows data of celebrities speaking English. The right block shows data of the same celebrities speaking German.}
    \label{fig:mav_celeb_v3}
\end{figure}

\begin{table}
\caption{Summary of MAV-Celeb dataset.}
\centering
 \scalebox{0.9}{
\begin{tabular}{lcc}
\toprule
\textbf{Dataset} & \textbf{English-Urdu} & \textbf{English-German} \\
\midrule
Languages & E/U/EU & E/G/EG \\
\# of celebrities & 70 & 58 \\
\# of male celebrities & 43 & 40 \\
\# of female celebrities & 27 & 18 \\
\# of videos & 560/406/966 & 212/216/428 \\
\# of hours & 59/32/91 & 8.2/6.6/14.8 \\
\# of utterances & 11835/6550/18385 & 2043/1769/3812 \\
Avg \# of videos per celebrity & 8/6/14 & 3.7/3.7/7.4 \\
Avg \# of utterances per celebrity & 169/94/263 & 35/31/66 \\
Avg length of utterances (in s) & 17.9/17.8/17.8 & 14.5/13.5/14.0 \\
\bottomrule
\end{tabular}
}
\label{tab:data_stats}
\vspace{-3mm}
\end{table}

\section{Grand Challenge description}

\noindent \textbf{Dataset.} Following the prior FAME 2024 Grand Challenge hosted at ACM Multimedia~\cite{saeed2024face,saeed2024synopsis}, we updated the dataset including an additional language as well as bilingual speakers, curating a new dataset split consisting of $58$ English-German speakers. 
The split provides language annotations, which allow to analyze the impact of multiple languages on face-voice association. As for the previous splits, the samples are 
obtained from YouTube videos, and consist of celebrities appearing in interviews, talk shows, and television debates~\cite{nawaz2021cross}. 
The visual data spans a vast range of setups, including different poses, motion blurs, background clutters, video qualities, occlusions, and lighting conditions. Moreover, since the videos originate from real-world situations, they reproduce the same challenges that are encountered when deploying face-voice association tools in real-world scenarios, such as noise, background chatter or music, overlapping voices, and compression artifacts. These aspects render the dataset both challenging for existing algorithms, and useful for developing algorithms that can have an impact on real applications. 
Fig.~\ref{fig:mav_celeb_v3} presents audio-visual samples from the newly collected dataset split, while Table~\ref{tab:data_stats} provides detailed statistics of the dataset.

The training and validation sets will be shared to the participants on the first day of the progress phase to allow the teams to develop their systems. In order to allow participants to benchmark their results, in the progress phase we will also release a pretrained model reaching competitive performance on the task. The progress phase will feature a leaderboard to keep participants up-to-date on the progress made by other teams on the validation set. The evaluation set (without ground-truth labels) will be shared on the first day of the evaluation phase. This will allow participants to test the generalization capability of the models developed and trained during the progress phase.

\begin{figure*}
\centering
\includegraphics[scale=0.75]{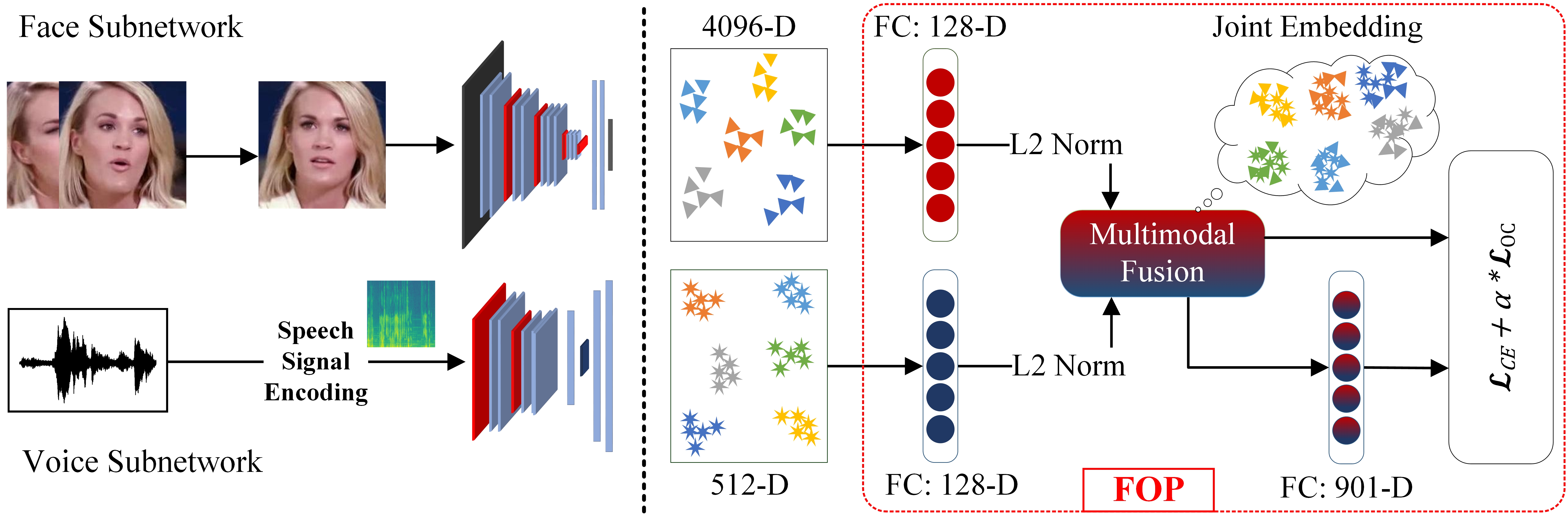}
   \caption{Overall architecture of the baseline method. 
   Face and voice embeddings are extracted by utilizing vision and audio encoders, respectively. Extracted features are then fed to linear layers to obtain the projected features of dimension $D$. Afterwards, embeddings are fused by using a gated feature fusion module. The fused features are fed to the logits layer. The model parameters are optimized by means of a linear combination of cross-entropy ($L_{CE}$) and orthogonal constraints ($L_{OC}$) losses. 
   } 
\label{fig:overall_framework_fusion}
\vspace{-2mm}
\end{figure*}

\noindent \textbf{Baseline Method \& Starter Kit.}
To allow participants to benchmark their results, we will release a pretrained instance of a competitive multimodal method for face-voice association task~\cite{saeed2022fusion}. 
The model consists of a two-branch network that takes as input the embeddings of faces and voices. The embeddings to be used as input to the face-encoding branch are obtained using a popular convolutional neural network pre-trained on a large-scale facial recognition dataset~\cite{parkhi2015deep}.  The embeddings to be used as input to the voice-encoding branch are obtained using an audio encoding network for speaker recognition~\cite{xie2019utterance} trained using the language available in the training set (i.e., the \textit{heard} language). The multimodal model further combines the face and voice embeddings, and is optimized by means of a loss function that imposes orthogonality constraints on the multimodal embeddings of different speakers (see Fig.~\ref{fig:overall_framework_fusion}).
We refer the readers to Saeed et al.~\cite{saeed2022fusion} and to the repository of the dataset\footnote{\href{https://github.com/mavceleb/mavceleb_baseline/tree/FAME-2025---Ahmed-A}{https://github.com/mavceleb/mavceleb\_baseline}} for more information on prior work on the baseline.

\noindent \textbf{Baseline results.} Table~\ref{tab:lang} provides the baseline face-voice association results  with the impact of multiple languages on both splits of MAV-Celeb. The FAME challenge $2026$ encourages participants to explore novel ideas to improve performance on heard and unheard languages.

\noindent \textbf{Challenge Setup.} 
The dataset consists of videos of several speakers. Each video is in one language only. However, each speaker is multilingual and appears in videos of at least two distinct languages. Each dataset set, i.e. the development set and the evaluation,  is divided into train and test splits following the so-called \textit{unseen-unheard} configuration~\cite{nagrani2018learnable}: the train and test splits consist of disjoint speakers speaking the same language~\cite{nagrani2018learnable}. Fig.~\ref{fig:verification+protocol} shows the evaluation protocol at train and test time. 
The evaluation is carried out on the task of cross-modal verification and on both a \textit{heard} language, i.e., available during training, and an \textit{unheard} language, i.e., \textit{not} available during training.
We refer to the dataset splits as V$1$-EU ($64$–$6$ train-test English-Urdu speakers), and V$3$-EG ($50$–$8$ train-test English-German speakers). V$1$-EU will be provided for the progress phase. The previous version of the dataset also includes V$1$-EU ($64$–$6$ train-test English-Urdu speakers)
Although we will not use V$2$-EH as dataset to track teams' progress on the competition leaderboard, we will also provide  V$2$-EH to allow teams to test their solutions on a separate dataset.
The final evaluation will be carried out on the test set of the newly-developed V$3$-EG split.
Alongside the audios (.wav) and images (.jpg), the datasets will include the .txt files of the face-voice pairs, for which each line has the following format:

\begin{itemize}
    \item \texttt{ysuvkz41 voices/English/00000.wav faces/English/00000.jpg} 
    \item \texttt{tog3zj45 voices/English/00001.wav faces/English/00001.jpg}
    \item \dots
    \item \texttt{ky5xfj1d voices/English/00002.wav faces/English/00002.jpg}
    \item \texttt{yx4nfa35 voices/English/01062.wav faces/English/01062.jpg}
\end{itemize}

The first entry of the line (e.g. ysuvkz41) represents the id of the pair. The remaining two represent the local path of the corresponding audio (e.g. voices/English/00000.wav) and of a visual frame extracted from the video (e.g. faces/English/00000.jpg). This .txt file is associated with the ground-truth .txt file, containing the id of the pair, and a 1 or a 0, depending on whether the face and voice correspond to the same speaker or not.

The dataset is publicly available and provided alongside pre-extracted features representing the audios and images as encoded with state-of-the-art pre-trained architectures. 
We release all information related to the challenge on the challenge website.\footnote{\href{https://mavceleb.github.io/dataset/competition.html}{https://mavceleb.github.io/dataset/competition.html}}

\noindent \textbf{Evaluation Metric.} As commonly done for tasks of face-voice association, we will evaluate the performance of participants' submissions with equal error rate (EER) as metric
. The EER is the value of the false acceptance rate (FAR) and false rejection rate (FRR) for which both the errors are equal; As for FAR and FRR, a low value of EER indicates a good performance of the system. We expect participants to submit a .txt file containing output scores for every pair in the test set, indicating the system's confidence that the face and voice are matching, or in other words, that they belong to the same person. As we will indicate in the challenge description, for a same .txt submission file, a face-voice test pair having a higher score than another face-voice test pair will be interpreted as the model having a higher confidence that the first pair is \textit{matching}, i.e., that the face and voice correspond to the same person, as compared to the second pair.   
As in previous editions of the challenge, we choose this evaluation setup for several reasons. First, EER does not require the confidence scores of different models to span the same ranges in order to compare the models' performance. Second, EER does not require the use of a fixed threshold, as opposed to other metrics such as precision. Furthermore, in real-world applications, system developers may optimize the threshold on the confidence score to convert it to a binary value that determines the model's prediction on whether the face and voice belong to the same or to different person(s), depending on their specific needs. With a high threshold, the FAR is expected to be low, while the FRR is expected to be high. In summary, to evaluate the performance of different systems, EER is more suitable than threshold-dependent metrics such as accuracy, since it is independent of the confidence score ranges and of the threshold.

\begin{table}
\footnotesize
\caption{Performance of baseline on cross-modal verification, between faces and voices across multiple language and on various test configurations of the MAV-Celeb dataset.
Results indicate EER (in \%), with lower values indicating a better performance. FOP~\cite{saeed2022fusion} is used as baseline method.}
\vspace{-2mm}
\begin{center}
\begin{tabular}{lccc}
\toprule
     \multicolumn{4}{c}{\textbf{V1-EU}}  \\
\midrule
Configuration & Eng. test  & Urdu test & Overall Score  \\
\midrule
 Eng. train     & 29.3  & 37.9 & \multirow{2}{*}{33.4}  \\
                                                 Urdu train     & 40.4  & 25.8 &  \\

\midrule 
\midrule
 \multicolumn{4}{c}{\textbf{V3-EG}} \\
\midrule
  & Eng. test  & German test & Overall Score   \\
\midrule
 Eng. train     & 34.5  & 43.7 & \multirow{2}{*}{40.2}  \\
                                                German train   & 43.2  & 39.6 &  \\

\bottomrule

\end{tabular}

\end{center}
\vspace{-4mm}
\label{tab:lang}
\end{table}

\noindent \textbf{Submission Format.} 
The grand challenge will be implemented using CodaBench\footnote{\url{https://www.codabench.org/competitions/9467/}}. Participants are expected to compute and submit text files including the \texttt{id} and confidence scores in the following format:
\begin{itemize}
\item  \texttt{ysuvkz41 0.9988}
\item  \texttt{tog3zj45 0.1146}
\item \dots
\item  \texttt{ky5xfj1d 0.6514}
\item  \texttt{yx4nfa35 1.5321}
\item  \texttt{bowsaf5e 1.6578}
\end{itemize}
Participants are expected to submit a zipped directory containing the submission files named as:

\begin{itemize}
\item \texttt{sub\_score\_English\_heard.txt}
\item \texttt{sub\_score\_English\_unheard.txt}
\item \texttt{sub\_score\_Urdu\_heard.txt}
\item \texttt{sub\_score\_Urdu\_unheard.txt}
\end{itemize}

In the progress phase, each team will be allowed to submit a maximum of $100$ submissions, with a maximum $10$ per day. In the evaluation phase, the number of total submission will be limited to $5$. The overall score will be computed as:
\begin{equation}
\text { Overall Score }=(\text {Sum of all EERs}) / 4
\end{equation}

\noindent \textbf{Rules for System Development.}
Since the FAME 2026 challenge aims at analyzing whether face-voice association capabilities translate across languages, we will enforce the following rules for participation:

\begin{itemize}
    \item A pretrained model on heard language is allowed. 
    \item A pretrained model on unheard language is \textit{not} allowed. The evaluation follows unheard-unseen and completely \textit{unheard} protocol. Each celebrity in the split has audio information in two languages; the model will be trained on one language (e.g., German) and then tested on both the heard language (e.g., German) and completely \textit{unheard} language (e.g., English).
    \item The participants are required to submit a $2$-page system description following the ICASSP template to the challenge organizers. Teams without system description will be disqualified from the challenge. Teams describing a setup that violates one of the above rules will be disqualified. 
    \item The participants are required to submit a link to a working version of their setup, e.g., on a platform for open-source development such as GitHub. Teams without code submission or with a setup that violates one of the above rules will be disqualified.
    \end{itemize}

\section{Registration Process}
The following Google Form will be used to allow participants to register their teams to the challenge.
\href{https://forms.gle/uNxC5Dmj3JVrCQi58}{Registration form}.

\section{Tentative Timeline}

\begin{itemize}
\item Registration Period: 15 Aug.~-- 15 Oct.~2025
\item Progress Phase: 1 Sep.~-- 31 Oct.~2025
\item Evaluation Phase: 1 Nov.~--~15 Nov.~2025
\item Challenge Results: 16 Nov.~2025
\item Submission of System Descriptions: 25 Nov.~2025
\item Challenge Paper Submission: 07 Dec.~2025
\end{itemize}

\section{Acknowledgments}
This research was funded in whole or in part by the Austrian Science Fund (FWF): \url{https://doi.org/10.55776/COE12} as part of the Cluster of Excellence \href{https://www.bilateral-ai.net/}{Bilateral Artificial Intelligence}.

\bibliographystyle{IEEEbib}
\bibliography{IEEEbib}

\end{document}